\documentclass[a4paper,twocolumn]{article}
\usepackage{gc2019eng_latex}

\newcommand{\norm}[1]{\left\lVert#1\right\rVert}
\graphicspath{{images/}}

\title{Predicting video saliency using crowdsourced mouse-tracking data}

\author{V.A.~Lyudvichenko$^1$, D.S.~Vatolin$^1$}

\email{vlyudvichenko@graphics.cs.msu.ru | dmitriy@graphics.cs.msu.ru}

\organization{$^1$Lomonosov Moscow State University, Moscow, Russia}

\abstract{%
This paper presents a new way of getting high-quality saliency maps for video, using a cheaper alternative to eye-tracking data.
We designed a mouse-contingent video viewing system which simulates the viewers' peripheral vision based on the position of the mouse cursor.
The system enables the use of mouse-tracking data recorded from an ordinary computer mouse as an alternative to real gaze fixations recorded by a more expensive eye-tracker.
We developed a crowdsourcing system that enables the collection of such mouse-tracking data at large scale.
Using the collected mouse-tracking data we showed that it can serve as an approximation of eye-tracking data.
Moreover, trying to increase the efficiency of collected mouse-tracking data we proposed a novel deep neural network algorithm that improves the quality of mouse-tracking saliency maps.
}


\authorsInfo
{
    Vitaliy Lyudvichenko is a Ph.D. student of Computer Graphics and Media Lab of Computer Science department of Lomonosov Moscow State University. 
    Dmitriy Vatolin is Vitaliy's scientific supervisor. 
}

\begin{document}

\maketitle

%\begin{multicols*}{2}

\section{Introduction}

When watching videos, humans distribute their attention unevenly.
Some objects in the video may attract more attention than the others.
This distribution can be represented by per-frame saliency maps defining the importance of each frame region for viewers.
The use of saliency can improve the quality of many video processing applications such as compression~\cite{Gitman2014} and retargeting~\cite{Lu2010} etc.

Therefore, many research efforts have been made to develop algorithms predicting saliency of images and videos~\cite{Borji2013}.
However, the quality of even the most advanced deep learning algorithms is insufficient for some video applications~\cite{Borji2018,SAVAM2}.
For example, deep video saliency algorithms slightly outperform eye-tracking data of a single observer~\cite{Lyudvichenko2019}, whereas at least 16 observers are required to get ground-truth saliency~\cite{Hollywood2_UCFSports_Mathe2015}.
% сказать как сложно предсказывать именно для видео

Another option to obtain high-quality saliency maps is to generate them from eye fixations of real humans using eye tracking.
Arbitrarily high quality can be achieved by adding more eye-tracking data from more observers.
However, collection of the data is costly and laborious because eye-trackers are expensive devices that are usually available only in special laboratories.
Therefore, the scale and speed of the data collection process is limited.

Eye-tracking data is not the only way to estimate humans' visual attention.
Recent works~\cite{SALICON,BubbleView} offered alternative methodologies to eye tracking that use mouse clicks or mouse movement data to approximate eye fixations on static images.
To collect such data a participant is shown an image on a screen.
Initially, the image is blurred, but a participant can click on any area of the image to see the original, sharp image in a small circular region around the mouse cursor.
This motivates observers to click on areas of images that are interesting to them.
Therefore, the coordinates of mouse clicks can approximate real eye fixations.

Of course, such cursor-tracking data of a single observer approximates visual-attention less effectively than eye-tracking data.
But in general, quality comparable with eye tracking can be achieved by adding more data recorded from more observers.
The main advantage of such cursor-based approaches is that they significantly simplify the process of getting high-quality saliency maps.
To collect the data only a consumer computer with a mouse is needed.
Thanks to crowdsourcing web-platforms like Amazon Mechanical Turk, the data can be collected remotely and at large scale.
It drastically speeds up the collection process and allows to increase the diversity of participants.

\begin{figure*}[ht]
    
    \begin{tabular}[h!]{p{0.47\textwidth}p{0.47\textwidth}}
    \centering{\includegraphics[width=\linewidth]{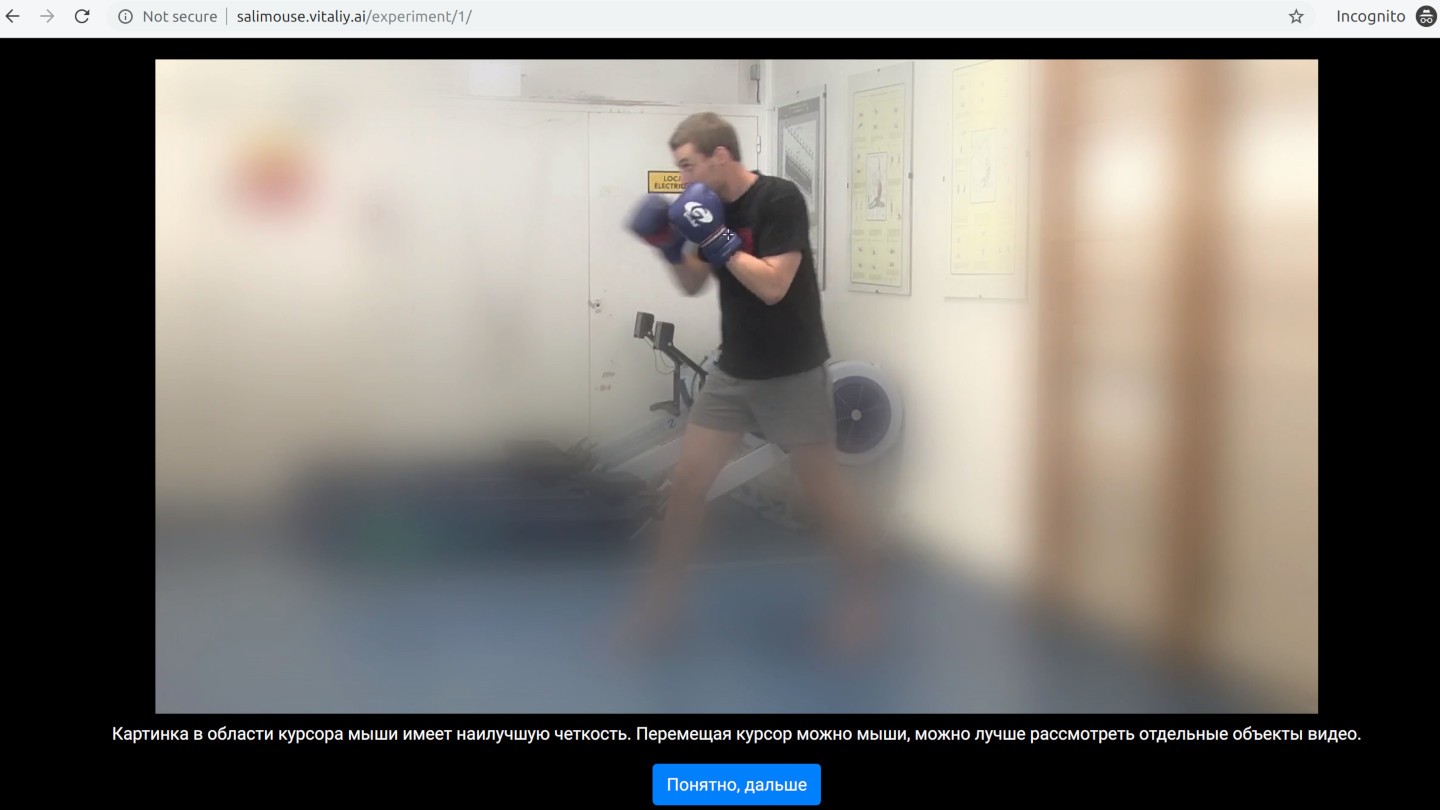}} &
    \centering{\includegraphics[width=\linewidth]{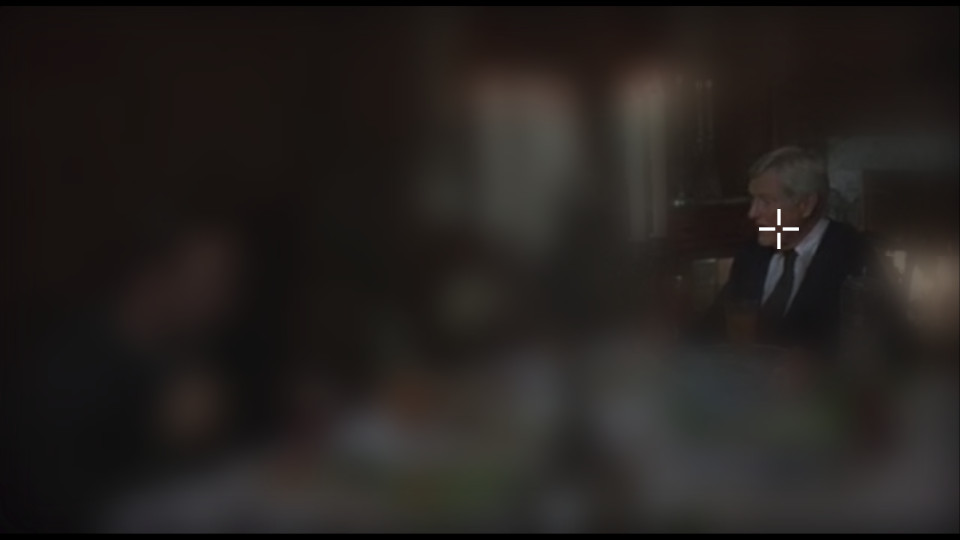}}
    \end{tabular}
    
    \vspace{-0.35cm}
    \caption{An example of a tutorial page and the mouse-contingent video player used in our system. The video around the cursor is sharp.}
    \vspace{-0.5cm}
    \label{fig:system-example}
\end{figure*}

In this work, we propose a cursor-based method for approximating saliency in videos and a crowdsourcing system for collecting such data.
To the best of our knowledge, it is the first attempt to construct saliency maps for video using mouse-tracking data.
We show participants a video which is being played in real time in the web-browser in a special video-player simulating the peripheral vision of the human visual system.
The player unevenly blurs the video in accordance with current mouse cursor position, the closer a pixel is to the cursor the less blur that is applied (Picture~\ref{fig:system-example}).
While watching the video a participant could freely move the cursor to see interesting objects without blurring.
Using the system we collected participants' mouse-tracking data who were hired on a crowdsourcing platform.
We performed an analysis of the collected data and showed that it can approximate eye-tracking saliency.
In particular, saliency maps generated from mouse-tracking data of two observers have the same quality as ones generated from eye-tracking data from a single observer.

However, cursor-based approaches, as well as eye-tracking, become less efficient in terms of added quality per observer when the number of observers goes up.
The contribution of each following observer to the overall quality is rapidly decreasing because the dependence between the number of observers and the quality is logarithmic in nature~\cite{Judd2012}.
Thereby, each following observer is more and more expensive in terms of cost per added quality.

%Authors~\cite{Gitman2014,SAVAM2} addressed this problem and proposed a semiautomatic approach for predicting saliency. 
To tackle this problem the semiautomatic paradigm for predicting saliency was proposed in~\cite{Gitman2014}.
Unlike conventional saliency models, semiautomatic approaches take eye-tracking saliency maps as an additional input and postprocess them which enables better saliency maps using less data.
%Instead of predicting saliency based on source video only, they postprocess saliency maps generated from eye-tracking data of single observer and produce saliency maps 
%The algorithm takes eye-tracking data of single observer as an additional input, applies a series of postprocessing algorithms on that data and manages to produce a saliency maps that have quality comparable with saliency maps directly generated from eye-tracking data of two observers.

We generalized the semiautomatic paradigm to mouse-tracking data and proposed a new deep neural network algorithm working within this paradigm.
The algorithm is based on SAM-ResNet~\cite{SAM-cornia2018} architecture, in which two modifications were made.
Since SAM-ResNet was designed to predict saliency in images, we firstly added an LSTM layer and adapted the SAM's attention module to exploit temporal cues of videos.
Then, we added a new external prior to the network which integrates mouse-tracking saliency maps into the network.
We showed that both modifications applied separately and jointly improve the quality.
In particular, we demonstrated that the algorithm can take mouse-tracking saliency maps that had the quality comparable with eye-tracking from three observers and improve them to the quality of eight observers.

%\end{multicols}
%\begin{multicols}{2}

\section{Related work}

The paper makes a contribution to two topics: cursor-based alternatives to eye tracking and semiautomatic saliency modeling.
Hereafter we provide a brief overview of these topics.

\textbf{Cursor-based alternatives to eye tracking.} 
There were many efforts to use mouse tracking as a cheap alternative to eye tracking. 
However, most of these efforts were focused on webpage analysis~\cite{Xu2016}. 
Therefore we provide an overview of the most notable universal approaches working with natural images.

Huang et al.~\cite{SALICON} designed a mouse-contingent paradigm that allowed the use a mouse instead of the eye tracker to record humans' behaviors of viewing static images.
They show participants the image for five seconds.
The shown image is adaptively blurred to simulate peripheral vision as though a participant's gaze is focused on the mouse cursor.
Participants can freely move the mouse cursor. 
Cursor coordinates are recorded, clustered and filtered to remove outliers.
Authors showed that such cursor-based fixations have high similarity with eye-tracking fixations.
Using AMT crowdsourcing platform they estimated saliency of 10000 images which were published as the SALICON dataset.

BubbleView~\cite{BubbleView} has a similar methodology, but it does not use the adaptive blurring and reveals the unblurred area of the image only when a participant clicks on it.
%They perform extensive study of their methodology on different types of images using different size of the circular reveal area.

Sidorov et al.~\cite{Sidorov2019} addressed the problem of temporal saliency of video, i.e. how a whole frame is important for viewers.
To estimate the temporal importance they show participants a blurred video and allow them to turn off blurring under the cursor when the mouse button is held down.
Participants have a limited amount of time when they can see unblurred frames, therefore they push the button down only on interesting frames.

To the best of our knowledge, our method is the first attempt to estimate spatial saliency of video using mouse-tracking data.

\textbf{Semiautomatic saliency modeling.} Lyudvichenko et al.~\cite{SAVAM2} proposed a semiautomatic visual-attention algorithm for video.
The algorithm takes eye-tracking saliency maps as an additional input and performs postprocessing transformations to them yielding saliency maps with better quality.
The postprocessing is done in three steps: firstly they propagate fixations from neighboring frames to the current frame according to motion vectors, then they apply brightness correction and add a center prior image to the saliency maps maximizing the similarity between the result and ground-truth.

We proposed a new semiautomatic deep neural network algorithm for mouse-tracking saliency maps which outperforms our previous algorithm~\cite{SAVAM2} if it uses mouse-tracking data.
%It outperforms the the previous algorithm~\cite{SAVAM2} because it learns more expressive transformations in end-to-end fashion.

\section{Cursor-based saliency for video}

We propose a methodology for high-quality visual-attention estimation based on mouse-tracking data and a system collecting such data using crowdsourcing platforms.
We show a participant the video in a special video player in real-time in full-screen mode.
The player simulates the peripheral vision of the human visual system by blurring the video as though the participant's gaze is focused on the mouse cursor.
The human eye retina consists of receptor cells, which are unevenly distributed throughout the eye, with a peak at the center of the field of view.
The central, foveal area is most clearly visible, whereas other, peripheral ones are blurrier.
We simulate that specificity by adaptively blurring video in accordance with the position of the mouse cursor.
A participant can freely move the cursor simulating shifting of the gaze.

To enable real-time rendering of the adaptively blurred frames we use a simple Gaussian pyramid with two layers $\mathbf{L}^0$ and $\mathbf{L}^1$, where $\mathbf{L}^0$ is the original frame, $\mathbf{L}^1$ is a blurred frame with $\sigma_1$.
The displayed image is constructed as follows: $\mathbf{I}_p = \mathbf{W}_p \mathbf{L}^0_p + (1 - \mathbf{W}_p) \mathbf{L}^1_p$, where $p$ is pixel coordinates and $\textbf{W}_p$ is a blending coefficient dependent on the retina density at $p$.
Thus, $\mathbf{W}_p = \exp{\left(-\nicefrac{\norm{p - g}^2}{2 \sigma_w^2}\right)}$, where $g$ is the position of the mouse cursor, $\sigma_w$ is a parameter.
Both parameters $\sigma_1$ and $\sigma_w$ represent the size of the foveal area and depend on screen size and the distance between the participant and the screen.
Since we record the data in uncontrolled conditions and cannot compute these parameters exactly we chose $\sigma_1 = 0.02w$ and $\sigma_w = 0.2w$, where $w$ is video width.

The system consists of front-end and back-end parts. 
The back-end part allocates videos among participants, stores the recorded data and communicates with a crowdsourcing platform.
Before watching videos the system shows three educational pages explaining how the video player works, Figure~\ref{fig:system-example} shows the first page.
The front-end part implements the video player using the HTML5 Canvas API.
Also, it checks that the participant's screen size is at least 1024 pixels width and its browser is able to render video at least 20 FPS.
We excluded data from participants who didn't pass these checks.

%\end{multicols}
\begin{figure*}[ht]
    \centering{\includegraphics[width=\linewidth]{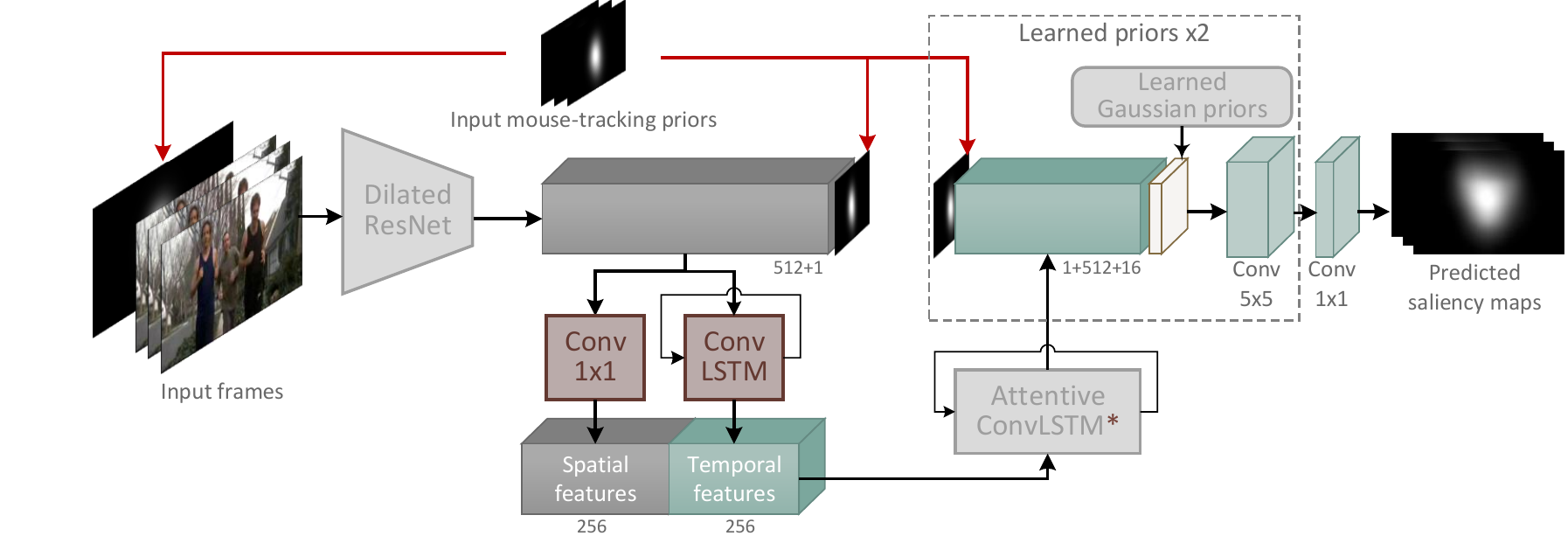}}
    
    \vspace{-0.3cm}
    \caption{%
    Overview of proposed temporal semiautomatic model based on SAM-ResNet~\cite{SAM-cornia2018}. %
    We introduce the external prior maps and concatenate them with the features of the input layer and three intermediate layers. %
    To make the network temporal-aware we introduce new spatiotemporal features and adapt the attentive ConvLSTM module so that it can pass the states to the following frames.
    The made modifications are marked by the red color on the schema.
    }
    \vspace{-0.4cm}
    \label{fig:dnn-scheme}
\end{figure*}
%\begin{multicols}{2}

\section{Semiautomatic deep neural network}

To improve saliency maps generated using the cursor positions as eye fixations we developed a new neural network algorithm.
The algorithm is based on SAM~\cite{SAM-cornia2018} architecture which was originally designed to predict saliency of static images.
Though SAM is a static model, its retrained ResNet version can outperform the latest temporal-aware models like ACL~\cite{ACL-wang2018} and OM-CNN~\cite{OM-CNN-Jiang2017,Lyudvichenko2019}.
Also, SAM architecture can be more easily adapted to video because its attentive module already uses LSTM layer to iteratively update the attention.

We make two modifications to the original SAM-ResNet architecture: adapt it for more effective video processing and add the external prior to integrate mouse-tracking saliency maps.
The modified architecture is shown in Figure~\ref{fig:dnn-scheme}.

Saliency models can significantly benefit from using temporal video cues.
Therefore we extract 256 temporal features in addition to 256 spatial features yielded from 2048 final features of ResNet subnetwork by 1$\times$1 convolution.
The temporal features are produced by additional convolutional LSTM layer with 3$\times$3 kernels which is fed with the final features of ResNet.
Spatial and temporal features are concatenated all together and passed to the Attentive ConvLSTM module.
Also, we make the Attentive ConvLSTM module truly temporal-aware by passing its states from the last iteration of the previous frame to the first iteration of the following frame.
It allowed reducing the number of per-frame iterations from 4 to 3 without quality loss.

Then we integrate the external map priors in three places of the network.
Firstly we add this prior to the existing Gaussian priors at the network head.
To learn more complex dependencies between the prior and spatiotemporal features we concatenate downsampled prior and the output of the ResNet subnetwork.
Also, we concatenate it with three RGB channels of source frames.
Since we use a pretrained ResNet network that expects the input with three channels, we update the weight of the first convolutional layer by adding a forth input feature initialized by zero weights.

\section{Experiments}

We used our cursor-based saliency system to collect mouse-movement data in 12 random videos from Hollywood-2 video saliency dataset~\cite{Hollywood2_UCFSports_Mathe2015} that are each 20--30 seconds long.
We hired participants on \href{http://www.subjectify.us/}{Subjectify.us} crowdsourcing platform, showed them 10 videos and paid them \$0.15 if they watched all videos.
In total, we collected data of 30 participants resulting in 22--30 views per video.

\begin{figure}[h]

% This file was created by matplotlib2tikz v0.7.4.
\begin{tikzpicture}

\definecolor{color6}{rgb}{0.890196078431372,0.466666666666667,0.76078431372549}
\definecolor{color3}{rgb}{0.83921568627451,0.152941176470588,0.156862745098039}
\definecolor{color5}{rgb}{0.549019607843137,0.337254901960784,0.294117647058824}
\definecolor{color1}{rgb}{1,0.498039215686275,0.0549019607843137}
\definecolor{color4}{rgb}{0.580392156862745,0.403921568627451,0.741176470588235}
\definecolor{color2}{rgb}{0.172549019607843,0.627450980392157,0.172549019607843}
\definecolor{color0}{rgb}{0.12156862745098,0.466666666666667,0.705882352941177}

\begin{axis}[
scale only axis,
width={0.87\linewidth},
height={0.87\linewidth},
legend cell align={left},
legend style={at={(1,0)}, anchor=south east, draw=white!70!black, fill opacity=0.5, text opacity=1},
tick align=outside,
tick pos=left,
x grid style={lightgray!92!black},
xlabel={\small Number of observers ($N$)},
xmajorgrids,
ymajorgrids,
xmin=1, xmax=8,
%xtick style={color=black},
y grid style={lightgray!92!black},
every minor grid/.style={solid, opacity=0.5},
every major grid/.style={solid, opacity=0.5},
tick align=inside,
ymin=0.38, ymax=0.74,
ytick style={color=black},
ytick={0.35,0.4,0.45,0.5,0.55,0.6,0.65,0.7,0.75},
yticklabels={0.35,0.40,0.45,0.50,0.55,0.60,0.65,0.70,0.75},
y tick label style = {rotate=35, anchor=east},
tick label style={font=\small}
]

\addplot [thick, color0, dashed]
table {%
1 0.728037681551539
10 0.728037681551539
};

\addplot [thick, color1, dashed]
table {%
1 0.687218857297055
10 0.687218857297055
};

\addplot [thick, color2, dashed]
table {%
1 0.678814656162539
10 0.678814656162539
};

\addplot [thick, color4, dashed]
table {%
1 0.659245192791033
10 0.659245192791033
};

\addplot [thick,  color5]
table {%
1 0.463659436926763
2 0.528820663183818
3 0.561778160477524
4 0.57384702124422
5 0.600520525042221
6 0.603366040012632
7 0.615378860280057
8 0.611185412720179
9 0.651212262742669
};

\addplot [thick, color3]
table {%
1 0.497617651138814
2 0.604622449685294
3 0.64632690234564
4 0.680344641701325
5 0.691476147339765
6 0.705701307169388
7 0.708681011871435
8 0.708377504922761
9 0.709249272938725
10 0.70377519037819
};

\addplot [thick, color6]
table {%
1 0.414249063582824
2 0.495441023281632
3 0.53930829558787
4 0.554636537004808
5 0.586421569907545
6 0.592131879623445
7 0.605244931294733
8 0.60483043505595
9 0.621360072734462
10 0.633490674001008
};

\addlegendentry{\scriptsize Temporal SAM with 10 MTO}
\addlegendentry{\scriptsize Static SAM with 10 MTO}
\addlegendentry{\scriptsize Temporal SAM without the prior}
\addlegendentry{\scriptsize Static SAM without the prior}
\addlegendentry{\scriptsize SAVAM~\cite{SAVAM2} with $N$ MTO}
\addlegendentry{\scriptsize $N$ eye-tracking observes}
\addlegendentry{\scriptsize $N$ MTO (mouse-tracking observers)}

% Alternative Y-axis
\node[anchor=north east, rotate=90, fill=white, fill opacity=0.5, text opacity=1] at (1,0.74) {\small Similarity Score~\cite{Judd2009}};

\end{axis}
\end{tikzpicture}

\vspace{-0.3cm}
\caption{ %
Objective evaluation of four configurations of our neural network: two semiautomatic versions using the prior maps generated from mouse-tracking data of 10 observers and two automatic versions without the prior maps. %
The networks are compared with the mean result of $N$ mouse- and eye-tracking observers as well as the SAVAM algorithm~\cite{SAVAM2} using $N$ mouse-tracking observers (MTO). %
Note, the number of observers is limited to half of the eye-tracking observers presented in the Hollywood-2 dataset~\cite{Hollywood2_UCFSports_Mathe2015}.}
\vspace{-0.49cm}

\label{fig:objective-evaluation}
\end{figure}
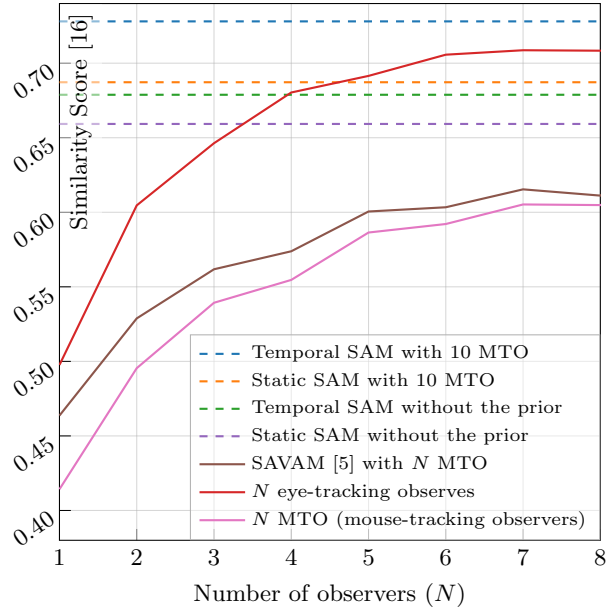

Using the collected data we estimated how good mouse- and eye-tracking fixations from the different number of observers approximate ground-truth saliency maps (generated from eye-tracking fixations).
Figure~\ref{fig:objective-evaluation} shows the results and illustrates that mouse-tracking of two observers have the same quality as eye-tracking of the single observer, so the data collected with the proposed system can approximate eye-tracking.

Note, when we estimated the eye-tracking performance of $N$ observers we compared them with the remaining $M-N$ observers of total $M$ observers. 
Therefore the eye-tracking curve has stopped increasing since $N=8$ because Hollywood-2 dataset has data of 16 observers only.
All our experiments convert fixation points to saliency maps using the formula $\textbf{SM}_p = \sum_{i=1..N}{\mathcal{N}(p, f_i, \sigma)}$, where $\textbf{SM}_p$ is the resulting saliency map value at pixel $p$, $f_i$ is the position of the $i$-th fixation point of $N$ and $\mathcal{N}$ is a Gaussian with $\sigma=0.0625w$, $w$ is video width.

We also tested how the previous semiautomatic algorithm~\cite{SAVAM2} works with mouse-tracking data from a different number of observers.
In the experiment, we computed the new optimal parameters for the algorithm.
Figure~\ref{fig:objective-evaluation} illustrates that the algorithm visibly improves only mouse-tracking saliency maps with few observers.
In particular, it improves mouse-tracking saliency maps of a single observer making them comparable with eye-tracking of a single observer.

Then we tested four configurations of proposed neural network architecture: two versions of the static variant and two versions of the temporal variant.
The static variant processes frames independently, whereas the temporal one uses temporal cues.
Each variant has the semiautomatic version using the external prior maps and the automatic version not using any external priors.
All architectures were trained on DHF1K~\cite{ACL-wang2018} and SAVAM~\cite{Gitman2014} datasets, the training set consisted of 297 videos with 86440 frames, the validation set contained 65 videos.
The NSS term was excluded from the original SAM's loss function since optimizing the NSS metric worsens all other saliency metrics.
All other optimization parameters are the same as those used in the original SAM-ResNet.

The static architecture variants were trained on every 25-th frame of the videos.
When training the temporal versions we composed minibatches from 3 consecutive frames of 5 different videos to use as large of a batch size as possible.
Also, we disabled training of batch normalization layers to avoid problems related to small batch size.

Since the collected mouse-tracking data wasn't enough for training the semiautomatic architectures we employed transfer learning technique and used eye-tracking saliency maps for the network's external prior.
The prior maps were eye-tracking saliency maps of 3 observers which have the same quality as mouse-tracking maps of 10 observers (according to Figure~\ref{fig:objective-evaluation}).

Figure~\ref{fig:objective-evaluation} shows the performance of all four trained networks where the external prior maps for the semiautomatic networks were generated from mouse-tracking data of 10 observers.
The figure demonstrates that the temporal configurations significantly outperform the static ones.
Thus, the added temporal cues improved the Similarity Score measure~\cite{Judd2009} of the original SAM~\cite{SAM-cornia2018} static version from 0.659 to 0.678, and the semiautomatic version from 0.687 to 0.728.

The semiautomatic versions improve their prior maps and have better quality than the automatic versions.
Also, they significantly outperform the semiautomatic algorithm proposed in~\cite{SAVAM2}.
It's worth noting that the best temporal semiautomatic configuration, which uses the prior maps generated from mouse-tracking data of 10 observers, outperforms eye-tracking of 8 observers. 
Since the prior maps have the same quality as 3 eye-tracking observers, the proposed semiautomatic algorithm actually improves saliency maps as though 5 more eye-tracking observers were added.

\section{Conclusion}

In this paper, we proposed a cheap way of getting high-quality saliency maps for video through the use of additional data.
We developed a novel system that shows viewers videos in a mouse-contingent video player and collects mouse-tracking data approximating real eye fixations.
We showed that mouse-tracking data can be used as an alternative to more expensive eye-tracking data.
Also, we proposed a new deep semiautomatic algorithm which significantly improves mouse-tracking saliency maps and outperforms traditional automatic algorithms. 

\section{Acknowledgments}

This work was partially supported by the Russian Foundation for Basic Research under Grant 19-01-00785 a.

\bibliographystyle{unsrt}
\bibliography{bibligraphy.bib}

\aboutAuthors

%\end{multicols*}

\end{document}